\documentclass[runningheads]{llncs}
\usepackage[T1]{fontenc}
\usepackage{graphicx}
\usepackage{color}
\usepackage{hyperref} 
\usepackage{subcaption}
\usepackage{xcolor}
\usepackage{adjustbox}

\begin{document}
\title{Spb3DTracker: A Robust LiDAR-Based Person Tracker for Noisy Environment}
% \inst{1}\orcidID{0000-1111-2222-3333}
\author{Eunsoo Im \and
Changhyun Jee \and
Jung Kwon Lee
% Eunsoo Im \and
% Changhyun Jee \and
% Jung Kwon Lee
}
\authorrunning{Im et al.}
% {Im et al.}
\institute{Superb-AI, 3D vision team\\
% Superb-AI Inc, 3D Vision Team \\
% \and
\email{eslim@superb-ai.com}
% \\
% \url{http://www.springer.com/gp/computer-science/lncs} 
% \and
}
\maketitle              % typeset the header of the contribution
\begin{abstract}
Person detection and tracking (PDT) has seen significant advancements with 2D camera-based systems in the autonomous vehicle field, leading to widespread adoption of these algorithms. However, growing privacy concerns have recently emerged as a major issue, prompting a shift towards LiDAR-based PDT as a viable alternative. Within this domain, "Tracking-by-Detection" (TBD) has become a prominent methodology. Despite its effectiveness, LiDAR-based PDT has not yet achieved the same level of performance as camera-based PDT.
This paper examines key components of the LiDAR-based PDT framework, including detection post-processing, data association, motion modeling, and lifecycle management. Building upon these insights, we introduce SpbTrack, a robust person tracker designed for diverse environments. Our method achieves superior performance on noisy datasets and state-of-the-art results on KITTI Dataset benchmarks and custom office indoor dataset among LiDAR-based trackers.

\end{abstract}
%
%
%
% https://ies0411.github.io/spb3DTracker/
% Project page at https://dk-liang.github.io/CLTR/.
\section{Introduction}
Lidar-based person detection and tracking (PDT) is crucial for ensuring safety and reliability in various domains, including autonomous driving, industrial safety, and crowd management \cite{Zhao2019}. While image-based trackers have been widely used, recent concerns about personal information protection have highlighted their limitations \cite{Ling2020}. As a result, LiDAR technology has emerged as a promising alternative, as LiDAR data does not contain personally identifiable information, making it impossible to distinguish specific individuals solely from this data \cite{Liu2018}.
However, LiDAR-based person tracking presents its own set of challenges. To address these, we delve into the tracking module and propose improved algorithms. LiDAR-based PDT generally fall into two categories: neural network-based approaches and tracking-by-detection (TBD) methods.
Neural network-based approaches, such as those utilizing Graph Neural Networks (GNNs), offer attractive end-to-end solutions \cite{Wang2019}. However, they often require ground truth (GT) labels for object IDs, which are scarce in real-world datasets and challenging to obtain \cite{Sattler2018}.
TBD has emerged as the predominant and most effective paradigm for LiDAR-based tracking in recent years \cite{Cai2021}. This paper focuses on enhancing the TBD mechanism due to its practicality and widespread applicability.
Traditional TBD tracking systems comprise several modules, including detection, association, motion modeling, and object management \cite{Caltagirone2020}. Each module requires careful analysis to adapt to person tracking scenarios. To address these challenges, we propose a comprehensive analysis of these modules, considering various conditions and scenarios.
Our research aims to improve the overall performance and reliability of LiDAR-based person tracking systems by optimizing each component of the TBD framework. This work contributes to the advancement of person tracking technology, which has significant implications for safety and efficiency in autonomous driving, industrial environments, and crowded public spaces \cite{Zhou2020}.

\begin{figure}[t]
\centering
\begin{subfigure}[b]{\columnwidth}
    \centering
    \includegraphics[width=0.7\columnwidth]{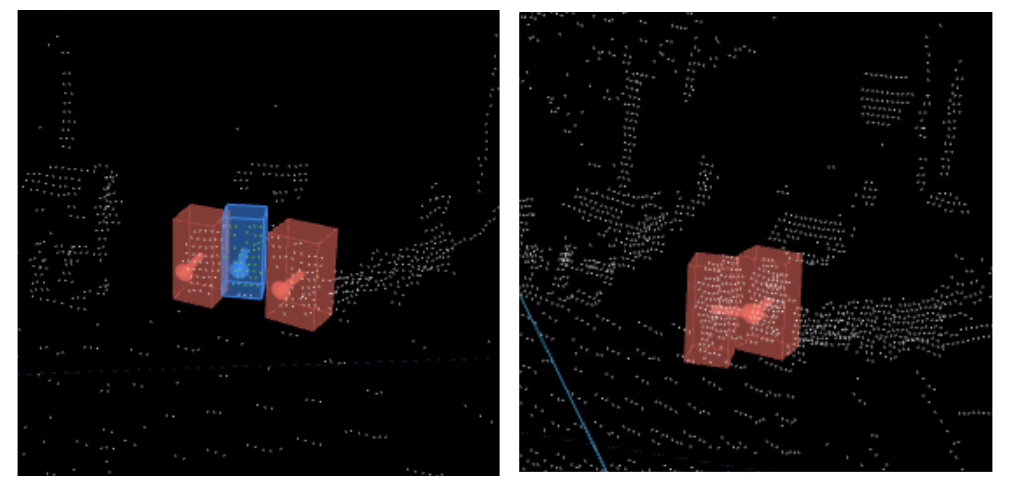}
    \caption{tracklet by associating high score 3D boxes}
    \label{f:intro_1}
\end{subfigure}
\vspace{0.5cm} % 그림 사이의 간격을 조정할 수 있습니다.
\begin{subfigure}[b]{\columnwidth}
    \centering
    \includegraphics[width=0.7\columnwidth]{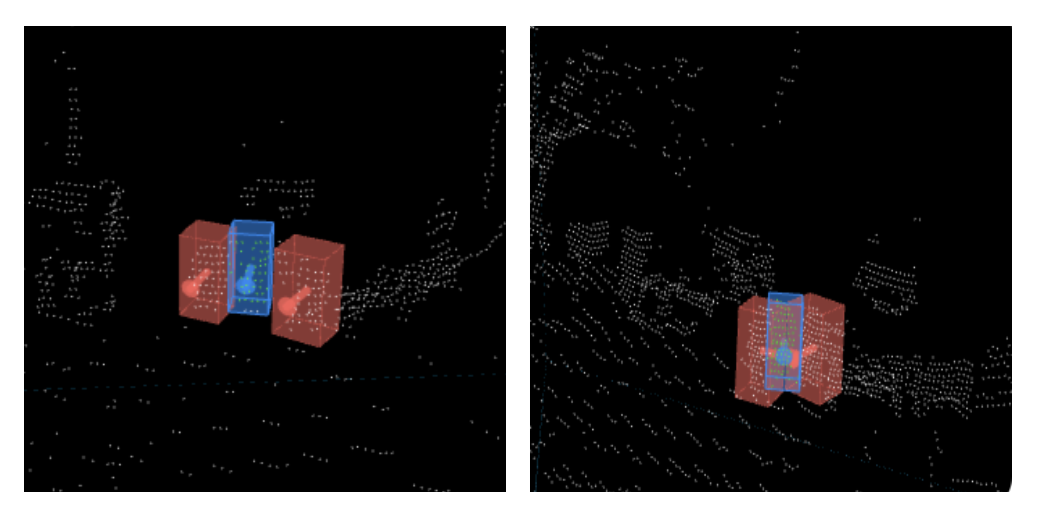}
    \caption{Tracklet by associating every 3D boxes}
    \label{f:intro_2}
\end{subfigure}
\end{figure}
In this paper, we propose several enhancements to the LiDAR-based PDT framework to improve trajectory robustness and continuity. Our approach addresses key limitations in current tracking-by-detection (TBD) methods for LiDAR data.
Firstly, we argue that relying solely on distance-based Intersection over Union (IoU) for association is insufficient \cite{Bewley2016}. We introduce a more robust approach that incorporates object similarity metrics and object dimensions \cite{Wu2019}. This enhancement improves the accuracy of object association across frames.
Secondly, we address the limitations of standard Kalman filters in handling non-linear motion and poor observation scenarios. person motion is often non-linear, and LiDAR detectors typically operate at 5Hz despite LiDAR sensors publishing raw data at 10Hz \cite{Voigtlaender2019}. The conventional Kalman filter, with its fixed system covariance, struggles to support non-linear motion and cope with inadequate observations \cite{Cho2014}. To overcome this, we propose an adaptive system covariance based on prediction confidence, allowing for more accurate state estimation in challenging scenarios \cite{Fritsch2019}.
Thirdly, we redesign the life-cycle management system. Many existing LiDAR-based tracker methods using TBD discard detection 3D boxes with low confidence scores \cite{Kim2021}. We challenge this approach, asserting that low-confidence 3D boxes may represent occluded objects. Consequently, our life-cycle management retains these detections to construct more complete trajectories efficiently \cite{Weng2019}. This approach improves the recall metric by preserving potential person tracks that might otherwise be discarded as "ghosts".
As illustrated in Figure \ref{f:intro_1} \ref{f:intro_2}, associating all 3D bounding boxes provides a more effective approach for identifying occluded objects. This method enhances the detection of objects that may be partially hidden or obstructed in the scene \cite{Sun2020}.
Fourthly, To demonstrate the generalization capability of our algorithm, we conduct experiments using both public datasets KITTI \cite{Geiger2013} and our own indoor office dataset. This indoor dataset is particularly valuable, as most public LiDAR datasets for person tracking focus on outdoor environments \cite{Behley2019}. By testing across diverse scenarios, we aim to evaluate the robustness of our tracking algorithm in different conditions.

\section{Related Works}

\subsection{2D Person Tracking}
Many recent person trackers are designed based on end-to-end deep learning networks. A common approach involves integrating recurrent layers with detector modules. For instance, ROLO combines the convolutional layers of YOLO with LSTM recurrent units \cite{Ning2017}, while TrackR-CNN extends multi-object tracking to include instance segmentation \cite{Voigtlaender2019}. Tractor++ offers an efficient tracking solution by utilizing bounding box regression to predict object positions in subsequent frames without requiring training on tracking data \cite{Bertasius2018}.
Other approaches focus on enhancing tracking performance through various techniques. Deep SORT integrates appearance information with the Simple Online and Realtime Tracking (SORT) technique, employing a recursive Kalman filter and frame-by-frame data association \cite{Wojke2017} \cite{zhang2022bytetrack}. This method incorporates an offline pre-training stage to learn a deep association metric using large-scale person re-identification datasets \cite{Leal-Taixe2016}. Similarly, some trackers adopt a single-stage approach, learning target detection and appearance embedding in a shared manner, often complemented by Kalman filters for location prediction \cite{Wang2020}.
While 2D trackers can leverage rich RGB information for appearance models, a resource not typically available in LiDAR-based 3D tracking, they often struggle with accurately representing 3D scene dynamics \cite{Bewley2016}. Conversely, 3D tracking methods can better handle scale variations and provide more accurate spatial information \cite{Geiger2013}. Ultimately, the design of person tracking methods should align with the specific characteristics of each modality to achieve optimal results, whether utilizing 2D image data, 3D LiDAR point clouds, or a fusion of multiple sensor inputs \cite{Weng2019}.

\subsection{Lidar-based Tracking}
Recent developments in Lidar-Based Tracking (LBT) have explored diverse strategies to enhance tracking performance in dynamic environments \cite{zhang2023lidar}. Some approaches focus solely on 3D bounding boxes detected from point clouds, offering rapid processing and comprehensive scene dynamics capture, albeit potentially underutilizing point cloud appearance features \cite{zhang2023lidar}.
Alternative methods have incorporated custom point cloud features or employed deep networks to estimate 3D bounding boxes from single-view images \cite{zhang2023lidar}\cite{pang2022simpletrack}. More recent approaches have shown promising results by combining features extracted from both RGB images and point clouds \cite{zhang2023strongfusionmot}.
Many LBT techniques rely on rule-based components. For instance, AB3DMOT, a widely-used baseline, employs Intersection over Union (IoU) for association and a Kalman filter for motion modeling \cite{zhang2023lidar}. Subsequent enhancements have focused on improving the association step, such as substituting IoU with Mahalanobis or L2 distance metrics \cite{zhang2023strongfusionmot}.
Researchers have also recognized the significance of lifecycle management in tracking systems \cite{zhang2023strongfusionmot}. Additionally, recent studies have investigated the application of graph neural networks (GNN) to address LBT in an end-to-end manner, with a focus on data association and active tracklet classification \cite{yan2017online}\cite{kim2022polarmot}.
As the field progresses, there is an increasing need for comprehensive studies on 3D Multi-Object Tracking (MOT) methods to identify areas for improvement and guide future research directions \cite{zhang2023strongfusionmot}. This systematic approach will help in addressing the challenges unique to LBT and advancing the state-of-the-art in this domain \cite{zhang2023lidar}\cite{zhang2023strongfusionmot}.

\subsection{Kalman Filter}
Kalman filters (KF) are widely used in tracking-by-detection methods for object tracking in autonomous driving due to their effectiveness in filtering noise from state estimations while handling noisy measurements \cite{zhang2023probabilistic}. However, standard KFs have limitations in real-world scenarios. They assume linear system models and Gaussian noise distribution, which may not hold in complex environments \cite{zhang2023probabilistic}\cite{zhang2020kalman}. They struggle with non-linear motion and poor observation conditions due to fixed system covariance, leading to error accumulation in state predictions, especially during periodic occlusions of objects \cite{zhang2020kalman}. Additionally, spatial distortions in 3D sensor detections can result in imprecise motion parameter estimations \cite{zhang2023probabilistic}.
To address these issues, advanced filtering techniques such as the Unscented Kalman Filter (UKF) and Cubature Kalman Filter (CKF) have been proposed \cite{zhang2020kalman}\cite{zhang20233d}. These filters are designed to better handle non-linearities and approximate state distributions more accurately than standard KFs \cite{zhang20233d}. Furthermore, adaptive filters can adjust system covariance dynamically to better respond to changing environments, improving tracking performance in complex, real-world autonomous driving scenarios \cite{zhang20233d}\cite{zhang2023karnet}\cite{lim2021pose}.

\begin{figure}[t]
\centering
\includegraphics[width=\linewidth]{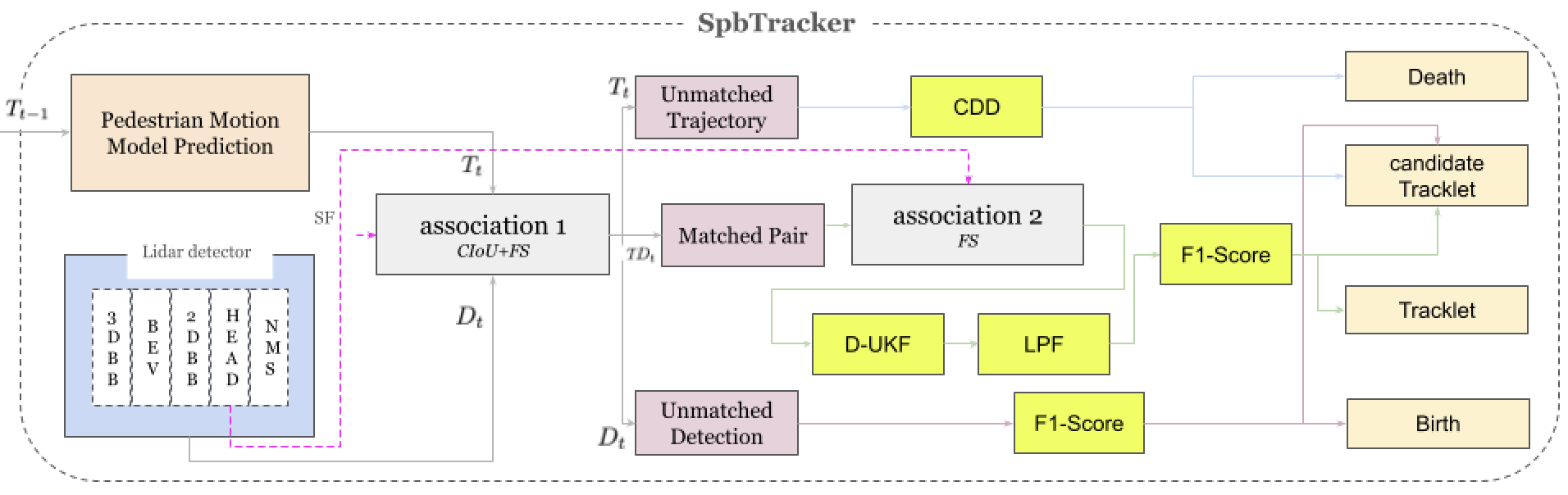}
\captionof{figure}{The SpbTracker pipeline associates results from 3D detector predictions with predictions from person motion models. The process involves three main cases:
1. Matched Pairs: Detections and trajectories that are initially matched undergo a two-stage association process. If they remain matched after this process, they are processed through D-UKF (Discrete Unscented Kalman Filter) and LPF (Low-Pass Filter) modules. Pairs with scores exceeding the F1-score threshold are then assigned to either active tracklets or candidate tracklets. 2. Unmatched Trajectories: These are processed through CDD (Confidence Decay and Deletion) modules. Their scores are then adjusted based on their distance from the ego vehicle. Trajectories with scores falling below the death threshold are either deleted or moved to the candidate tracklet pool. 3. Unmatched Detections: Among the unmatched detections, those with scores above the F1-score threshold are added to the birth pool or the candidate tracklet pool.}
\label{Fig:overview}
\end{figure}

\section{SpbTracker}
\subsection{Detection Process}
\begin{figure}[htbp]
\centering
\includegraphics[width=0.8\columnwidth]{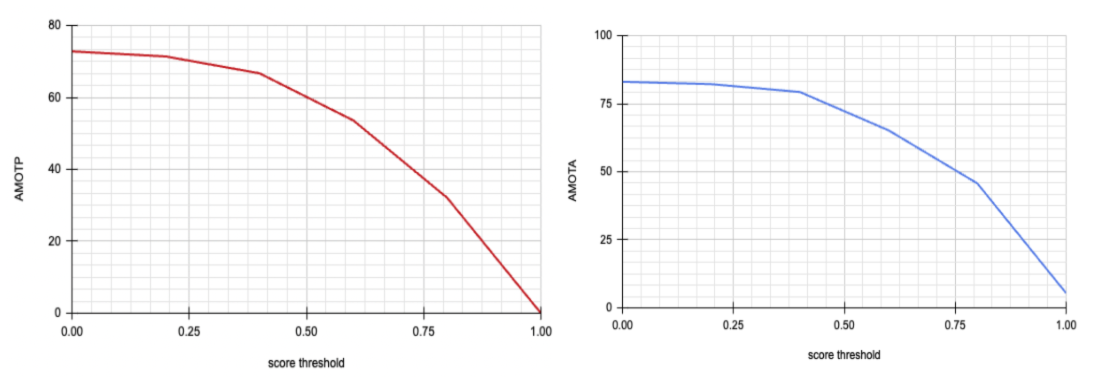}
\caption{The graphs illustrate the relationship between correlation score threshold and two metrics: AMOTP (left) and AMOTA (right) of pedestrian. As the threshold increases, both metrics decrease, indicating a decline in tracking performance. This ablation study was conducted using the DSVT detector model on the KITTI validation dataset. The results highlight the importance of carefully selecting the correlation score threshold to optimize tracking accuracy and precision.}
\label{f:confidence_threshold}
\end{figure}

Our approach diverges from conventional algorithms that typically retain only high-confidence 3D detection boxes. Instead, we implement a more inclusive strategy that avoids strict confidence thresholds \cite{zhang2023probabilistic}. This method aims to preserve potential tracklets and enhance recall performance.
In our pipeline, we employ the DSVT (Dynamic Sparse Voxel Transformer) detection model\cite{wang2023dsvt} and apply Non-Maximum Suppression (NMS) to refine the detection results. Unlike image-based approaches, our LiDAR-based method focuses on 3D space, simplifying the post-processing steps.
By eschewing a fixed confidence threshold, our approach seeks to strike a balance between retaining potentially valid detections and mitigating false positives. This strategy is particularly effective in challenging scenarios where low-confidence detections may still represent genuine objects.
As illustrated in Fig. \ref{f:confidence_threshold}, our experiments reveal an inverse relationship between the confidence threshold and tracking performance metrics (AMOTP and AMOTA). Lower thresholds correspond to improved metric scores, underscoring the importance of managing "ghost" objects (false positives) in enhancing recall.
\begin{table}[h]
% \centering
\caption{Comparison of person AP for detectors trained from scratch and from pretrained models, with and without shuffling. The experiments use the KITTI validation dataset and our custom office dataset.}
\begin{center}
\label{t:pretrained_model}
\renewcommand{\arraystretch}{1.4} 
\begin{tabular}{cccc}
\hline
\textbf{Method}                                                                     & \textbf{KITTI {[}AP{]}$\uparrow$} & \textbf{Office{[}AP{]}$\uparrow$} \\ \hline
From scratch                                                                        & 49.32                      & 64.12                               \\
\begin{tabular}[c]{@{}c@{}}From pretrained model (w/o batch shuffle)\end{tabular} & 48.32                      & 64.21                        \\
\begin{tabular}[c]{@{}c@{}}From pretrained model (batch shuffle)\end{tabular}     & 54.92                      & 75.89                               \\ \hline
\end{tabular}

\end{center}

\end{table}

Given the limited availability of person data in the KITTI dataset\cite{geiger2013vision}, we adopted a strategy to enhance our model's performance through transfer learning and multi-dataset training. This approach is particularly effective for image-based detection tasks, where pre-trained models often yield superior results.
To create a robust baseline model, we utilized a combination of another public datasets including person class. Rather than training sequentially on each dataset, we implemented a batch-wise shuffling technique. This method ensures that during each training iteration, the model is exposed to a diverse range of samples from all datasets simultaneously.
Our training procedure can be summarized as follows:
\begin{itemize}
\item We combined samples from KITTI, our custom dataset, and another public dataset.
\item During batch formation, we randomly shuffled samples from all datasets.
\item This shuffled batch was then used for training, allowing the model to learn from diverse data sources concurrently.
\end{itemize}

After creating this pre-trained model using the multi-dataset approach, we fine-tuned it exclusively on the KITTI dataset. This two-stage process allows our model to benefit from the rich, diverse data of multiple sources while still specializing in the specific characteristics of the KITTI dataset.
This methodology aims to address the scarcity of person data in KITTI by leveraging additional datasets, potentially improving the model's generalization capabilities and performance on the target dataset.
Table \ref{t:pretrained_model} demonstrates that using a pretrained model with shuffling improves the Average Precision (AP).

\subsection{Motion Model}
Most previous methods apply simplistic motion models that fail to adequately account for the complexity of a person's movement. persons can move freely along the X and Y axes and rotate around the Z-axis (yaw direction) according to the right-hand rule. Furthermore, analysis of person detection results reveals the difficulty in distinguishing between front and back views of individuals.
Consequently, prediction models relying solely on heading angle may be inaccurate. To address these limitations, we propose formulating the state of an object trajectory as an 11-dimensional vector:
\begin{equation}
[x, y, z, \theta, x_{v}, y_{v}, x_{a}, y_{a}, w, l, h]
\end{equation}
$(x, y, z)$ represents the 3D location of the object's geometric center.$(\theta)$ represents the object's orientation around the Z-axis.$(v_x, v_y)$ represents the object's velocity components in the X and Y directions.$(a_x, a_y)$ represents the object's acceleration components in the X and Y directions.$(w, l, h)$ represents the object's 3D dimensions (width, length, height)

\subsection{Dynamic Unscented Kalman Filter}
The standard Kalman Filter is widely used in Track-By-Detection (TBD) tracking modules for associating motion model systems with measurement systems (detection systems). Most previous methods assume linear object motion and Gaussian distribution. However, person motion is inherently non-linear, and the typical LiDAR sensor frame rate of 10Hz, coupled with TBD detection module latency of 0.3-0.7 seconds, often violates the Gaussian distribution assumption in the tracking module.
To address these issues, we propose a Dynamic Unscented Kalman Filter (D-UKF). We leverage the D-UKF to estimate the trajectory state. The prediction process can be described by:
\begin{equation}
(\chi_{t-1},W_{t})= (T_{t-1},P_{t-1},\kappa)
\end{equation}
\begin{equation}
(T_{t}, P_{t})=UT(f(\chi_{t-1}),W_{t},Q)
\end{equation}
\begin{equation}
(D_{t}, P_{z,t})=UT(h(\chi_{t-1}),W_{t},R)
\end{equation}
where $T$ denotes the prediction model, $D$ means the detection model, $P$ is the covariance matrix of the previous frame, $Q$ is the prediction model system's noise, and $R$ is the detection model system's noise. $f(\cdot)$ represents the motion model, and $h(\cdot)$ is the detection model.
The motion model is defined as:
\begin{equation}
x_{t} = x_{t-1} + v_{x,t-1}\Delta t + \frac{1}{2}a_{x,t-1}\Delta t^{2}
\end{equation}
\begin{equation}
y_{t} = y_{t-1} + v_{y,t-1}\Delta t + \frac{1}{2}a_{y,t-1}\Delta t^{2}
\end{equation}

\begin{table}[t]
\caption{Performance Comparison of Filters Using 10Hz Data}
\label{t:filter_1}
\centering
\renewcommand{\arraystretch}{1.4} 
% \resizebox{\columnwidth}{!}{
\begin{tabular}{lllll}
\hline
\textbf{Dataset}    & \textbf{Method} & \textbf{sAMOTA$\uparrow$} & \textbf{AMOTA$\uparrow$} & \textbf{IDs$\downarrow$} \\ \hline
KITTI & KF              & 99.12           & 91.02          & 9          \\
KITTI & UKF             & 99.28           & 92.19          & 8          \\
KITTI & D-UKF           & 99.27           & 92.12          & 8          \\ \hline

Office & KF              & 89.01           & 85.33          & 55         \\
Office & UKF             & 89.21           & 85.52          & 52         \\
Office & D-UKF           & 89.12           & 85.48          & 52          \\ \hline

\end{tabular}
% }
\end{table}
\begin{table}[t]
\caption{Performance Comparison of Filters Using 5Hz Data}
\label{t:filter_2}
\centering
\renewcommand{\arraystretch}{1.4} 
% \resizebox{\columnwidth}{!}{
\begin{tabular}{lllll}
\hline
\textbf{Dataset} & \textbf{Method} & \textbf{sAMOTA$\uparrow$} & \textbf{AMOTA$\uparrow$} & \textbf{IDs$\downarrow$} \\ \hline
KITTI & KF              & 94.24           & 88.88          & 25          \\
KITTI & UKF             & 96.88           & 89.12          & 15          \\
KITTI & D-UKF           & 97.15           & 91.77          & 13          \\ \hline

Office & KF              & 77.98           & 71.23          & 104          \\
Office & UKF             & 80.12           & 76.12          & 99          \\
Office & D-UKF           & 82.98           & 78.56          & 85          \\ \hline
\end{tabular}
% }
\end{table}

Traditional Kalman Filters use fixed measurement system covariances $R$. In an ideal scenario, this error would be zero. However, in real-world applications, this is not achievable. Our D-UKF approach, when generating sigma points, spreads these points according to the system covariance.
Unlike traditional KF that use fixed system noise, our SpbTracker dynamically adjusts system noise based on confidence of detection. We propose an D-UKF that modifies the measurement system covariance as follows:

\begin{equation}
 S_{k}=\sum_{2L}^{i=0}W_{i}[\gamma^{(i)}_{k|k-1}-\hat{z_{k|k-1}}][\gamma^{(i)}_{k|k-1}-\hat{z_{k|k-1}}]^{T}+R_{init}
\end{equation}
\begin{equation}
\nu_{k}=z_{k}-\hat{z_{k|k-1}}
\end{equation}
\begin{equation}
R_{k}=\frac{1}{confidence}[(1-\alpha )R_{k-1}+\alpha (\nu_{k} \nu_{k}^{T}-S_{k})]
\end{equation}
% \begin{equation}
% R_{k}=(1-\alpha )R_{k-1} + \alpha (\nu_{k}\nu_{k}^{T}-S_{k} ) ,  \nu_{k}=z_{k}-\hat{z}_{k|k-1}
% \end{equation}
% \begin{equation}
% R_{init} = \frac{1}{confidence} R_{baseline}
% \end{equation}
This adaptive mechanism allows the filter to adjust to varying levels of measurement uncertainty, improving tracking performance across diverse scenarios without manual parameter adjustments. By incorporating this approach, our method aims to enhance robustness and generalizability in PDT, particularly in challenging environments with complex motion patterns and varying sensor characteristics.
Tables \ref{t:filter_1} and \ref{t:filter_2} demonstrate the effectiveness of our Dynamic UKF in noisy environments. We conducted an ablation study using the KITTI validation dataset. For the 5Hz dataset, we analyzed every other data point in the sequence to maintain consistency.

\subsection{Association}
The Generalized 3D Generalized IoU (GIoU) is a widely adopted association metric in object tracking. This metric enhances the traditional IoU (Intersection over Union) by accounting for the spatial relationship between non-overlapping bounding boxes, thereby providing a more comprehensive measure of object overlap. However, GIoU solely reflects the proportion of overlap, neglecting crucial factors such as volume size and object viewpoint. To address these limitations, we incorporate the concept of Complete-IoU (CIoU) into our association framework. CIoU extends the capabilities of GIoU by considering additional geometric properties, thus offering a more nuanced approach to object association.
We suggest new metric by modifying CIoU metric as Modified Complete-IoU(MCIoU). MCIoU is specifically designed to reflect the unique characteristics of person tracking, particularly taking into account the ratio of person height:

\begin{equation}
v = \frac{4.0}{\pi}(arctan\frac{h_{s}}{Area_{s}}-arctan\frac{h_{t}}{Area_{t}})
\end{equation}
\begin{equation}
\alpha = v (\frac{v}{1-GIoU}+1)
\end{equation}
\begin{equation}
MCIoU = GIoU + \alpha
\end{equation}
Nevertheless, IoU-based association methods inherently face challenges in Re-Identification (Re-ID) tasks, particularly when dealing with objects that have been absent from the scene for extended periods. This limitation arises from the reliance on threshold distances for matching, which can fail to associate objects that have undergone significant spatial displacement over time. To mitigate this issue, we augment our association strategy by incorporating feature similarity measures. This additional dimension allows for more robust matching, especially in scenarios where spatial proximity alone is insufficient.
we utilize bird's eye view(BEV) features sampled via region of interest(ROI) as feature similarity. Considering that the matched object ROI features of the two branches, we design the feature similarity(FS) as:
\begin{equation}
FS = exp\left ( \frac{f_{s}\left ( f_{t} \right )^{T}}{\left\| f_{s} \right\|_{2} \left\|f_{t} \right\|_{2}} \right )
\label{eq:feature_similarity}
\end{equation}
\begin{figure}[h]
\centering
\includegraphics[width=0.68\columnwidth]{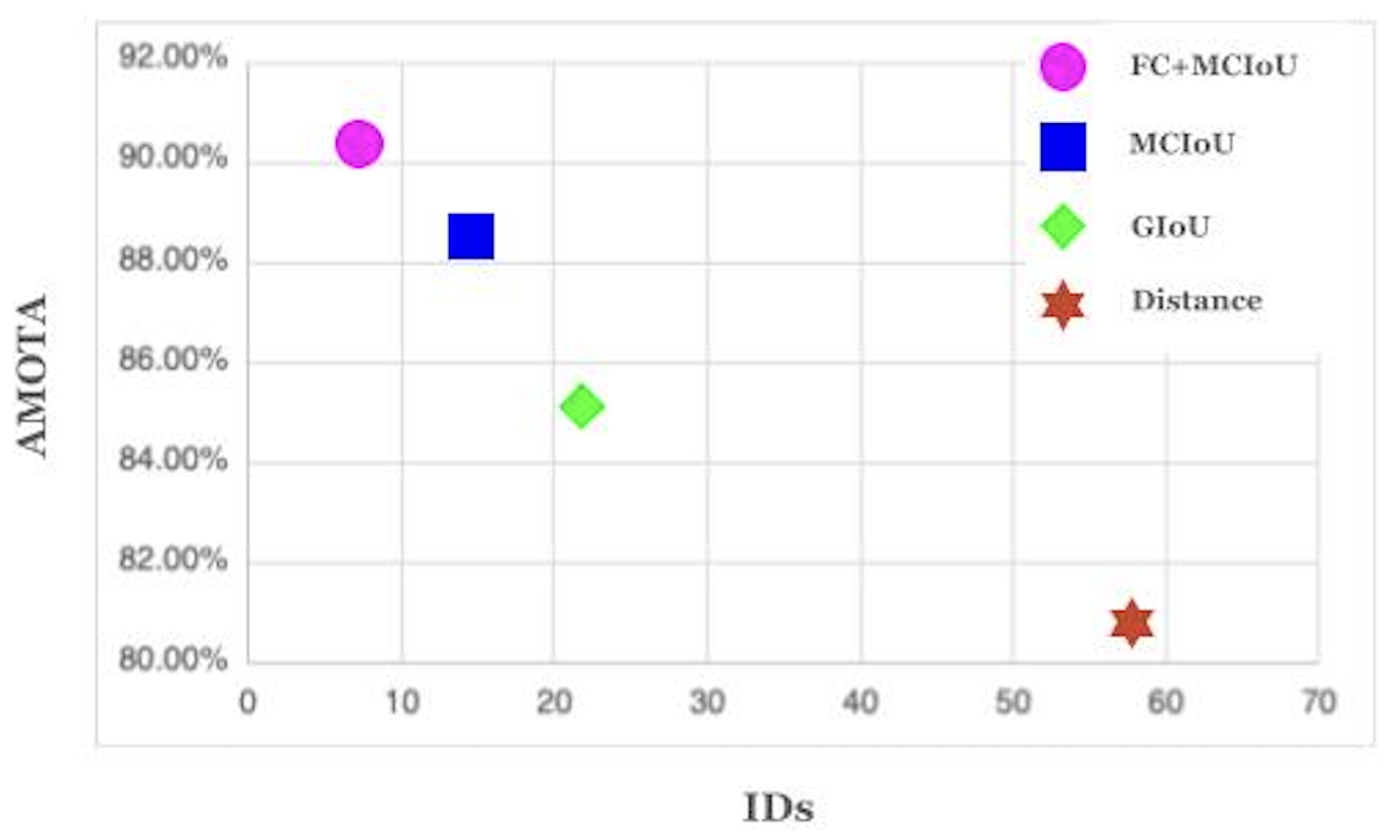}
\caption{Comparison of association metrics}
\label{f:association}
\end{figure}
Our final association metric is derived from a weighted sum of the MCIoU score and FS measure. This composite approach leverages the strengths of both geometric and appearance-based cues, resulting in a more robust and versatile association framework capable of handling diverse tracking scenarios.
\begin{equation}
\omega MCIoU + (1-\omega) FS, (0<\omega<1)
\label{eq:weight_sum}
\end{equation}
Fig. \ref{f:association} presents a comparison of association metrics on the KITTI validation set. The optimal method is positioned closest to the bottom-left corner, indicating the lowest number of ID-Switches and the highest AMOTA score. Our proposed metric demonstrates superior performance among the compared methods, as evidenced by the plot.

\subsection{Life Cycle Management}
While previous approaches often rely exclusively on high-confidence 3D bounding boxes, our method incorporates all predicted 3D boxes to enhance recall metrics. To address the challenges associated with this comprehensive approach, we adopt a holistic strategy as follows:
Firstly, we utilize the F1 score as the classification criterion. We associate the track pool with all detection results using our previously described association method. This approach, however, can potentially lead to decreased precision due to the inclusion of low-confidence detections and increased computational complexity in the association step, which employs the Hungarian algorithm.
In contrast to conventional algorithms that use distance-based thresholds for association, we employ a Feature Similarity (FS) metric. Among pairs exceeding a distance threshold, we compare their FS values. Pairs with FS values above a predetermined threshold are retained in the pair memory pool.
Matched pairs undergo processing through a Dynamic Unscented Kalman Filter (UKF) and a Low-Pass Filter (LPF) for confidence score smoothing. The LPF mitigates rapid fluctuations in confidence scores, applying to both matched trajectory scores and detection result scores. If the LPF result exceeds the F1-score threshold, the corresponding tracklet is classified as active.
The LPF is defined as:
\begin{equation}
LPF = \omega T_{score} + (1-\omega)D_{score}, (0<\omega<1)
\label{eq:LPF}
\end{equation}

\begin{figure}[t]
\centering
\includegraphics[width=\columnwidth]{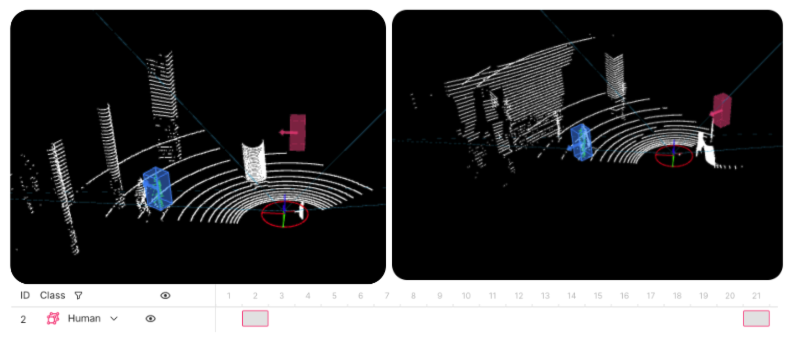}
\caption{Long-term tracking persistence}
\label{f:memory}
\end{figure}

where $T_{score}$ represents the matched trajectory confidence score, and $D_{score}$ denotes the matched detection result confidence score.
For unmatched detection results, if the score exceeds the F1-score threshold, the detection is either initialized as a new tracklet or classified as a candidate trajectory.
Unmatched trajectories undergo score decay based on their distance from the ego-vehicle(or lidar), reflecting the increased likelihood of object disappearance at greater distances. Trajectories with scores falling below the death threshold are removed from the manager memory.
The Confidence Decay Distance (CDD) is calculated as:
\begin{equation}
CDD = \frac{\sqrt{(x_{t}-x_{ego})^{2}+(y_{t}-y_{ego})^{2}+(z_{t}-z_{ego})^{2}}}{MAX\:RANGE}
\label{eq:CDD}
\end{equation}
Fig. \ref{f:memory} demonstrates the long-term tracking persistence capability of our life-cycle memory approach. In frame 2, SpbTracker successfully tracks a person labeled as ID-2. From frames 3 to 20, the object becomes occluded and temporarily disappears from view. Despite this extended period of occlusion, SpbTracker maintains the object's identity in memory. When the object reappears in frame 21, SpbTracker correctly recognizes it as ID-2, avoiding an identity switch (IDsw).
This description effectively conveys the key aspects of your tracking system's performance, highlighting its ability to maintain object identity through extended periods of occlusion.

\section{Experiments}

\subsection{Experimental Setup}
We implemented our proposed methodology using Python and C++, and conducted experiments on an Intel Xeon Gold 5218R CPU (2.10GHz) with 125GB RAM. All reported results reflect online performance.
\subsubsection{Datasets}
We evaluated our method on two datasets: the KITTI tracking dataset, and a our own dataset.
The KITTI tracking dataset was recorded in Karlsruhe, Germany, using a 64-beam LiDAR sensor at a 10Hz frame rate. Following the convention in , we used sequences 1, 6, 8, 10, 12, 13, 14, 15, 16, 18, and 19 as the validation set.
Our custom Office dataset was captured using a Hesai QT128 LiDAR sensor. This sensor features a vertical Field of View (FOV) of 105.2° [-52.6°, +52.6°], a horizontal FOV of 360°, a range of 0.05 to 50 meters, and operates at a frame rate of 10Hz. We collected person tracking data in an indoor environment using this LiDAR sensor.

\begin{table}[t]
\centering
\caption{A comparison of existing algorithms for the KITTI 2D Pedestrian Tracking task on the test set is presented, using the 2D Pedestrian Tracking Metric. In the results table, the best performance for each metric is highlighted in red, while the second-best performance is marked in blue.}
\label{t:test-benchmark}
\renewcommand{\arraystretch}{1.4} 
\begin{adjustbox}{width=\textwidth} 
\begin{tabular}{l c c c r r r r r}
% \begin{tabular*}{@{\extracolsep{\fill}}l c c c r r r r r@{}}
\hline
\textbf{Model} & \textbf{Paper} & \textbf{Input} & \textbf{Method} & \textbf{sMOTA$\uparrow$}               & \textbf{MOTA$\uparrow$}                & \textbf{MOTP$\uparrow$}                & \textbf{HOTA$\uparrow$}                & \textbf{IDSW$\downarrow$}              \\ \hline
MSAMOT         & Sensors '22     & 2D + 3D        & TBD             & 24.16                        & 47.86                        & 64.35                        & {\color[HTML]{3166FF} 44.73}                        & {\color[HTML]{3166FF} 209} \\
FNC2           & TIV '24         & 2D + 3D        & TBD             & {\color[HTML]{3166FF} 30.86} & {\color[HTML]{FE0000} 56.05} & {\color[HTML]{FE0000} 65.68} & {\color[HTML]{FE0000} 46.55} & 335                        \\
EagerMOT       & ICRA '21        & 2D + 3D        & TBD             & 28.01                        & 49.82                        & 64.42                        & 39.38                        & 496                        \\
StrongFusion   & Sensors '22     & 2D + 3D        & TBD             & 16.54                        & 39.04                        & 63.89                        & 43.42                        & 316                        \\
JRMOT          & IROS '20        & 2D + 3D        & TBD             & 29.78                        & 45.31                        & 72.22                        & 34.24                        & 631                        \\ \hline
AB3MOT         & IROS '20        & 3D             & TBD             & 20.80                        & 38.13                        & 64.54                        & 37.81                        & 879                        \\
PolarMOT       & ECCV '22        & 3D             & GNN             & 24.61                        & 46.98                        & 64.59                        & 43.59                        & 270                        \\ \hline
\textbf{Ours}  & -              & 3D             & TBD             & {\color[HTML]{FE0000} 32.72} & {\color[HTML]{3166FF} 53.55} & {\color[HTML]{3166FF} 65.28} & 43.25                        & {\color[HTML]{FE0000} 200} \\ \hline
\end{tabular}
\end{adjustbox}
\end{table}

\subsubsection{Evaluation Metrics and Criteria}
For the KITTI dataset, we conducted 3D MOT evaluation on the validation set, as the test set only supports 2D MOT evaluation and its ground truth is not publicly available. We followed the KITTI convention, presenting results for Pedestrian.
Regarding matching criteria, we adopted the convention from the KITTI 3D object detection benchmark, using 3D IoU to determine successful matches. Specifically, we employed 3D IoU thresholds (IoU thres) of 0.25 for pedestrian.

On KITTI,
we also report sAMOTA [56] and HOTA [31] metrics, which allows us to analyze
detection and association errors separately. We also report the number of identity
switches (IDs) at the best performing recall.

\subsection{Benchmark results}
\begin{table}[t]
\centering
\renewcommand{\arraystretch}{1.4} 
\caption{KITTI 3D Pedestrian validation set benchmark}
\label{t:kitti_val}
\begin{tabular}{llllll}
\hline
\textbf{Model}    & \textbf{Input} & \textbf{sAMOTA}$\uparrow$ & \textbf{MOTA}$\uparrow$  & \textbf{recall}$\uparrow$  & \textbf{IDs}$\downarrow$ \\ \hline
EagerMOT & 2D+3D & 92.95  & 93.14 & 93.61 & 36  \\
PolarMOT & 3D    & 94.08  & 93.48 & 93.66 & 9   \\
AB3MOT   & 3D    & 73.18  & 66.98 & 72.82 & {\color[HTML]{FE0000} 1}   \\
AB3MOT*  & 3D    & 88.56  & 82.14 & 88.13 & 87   \\ \hline
\textbf{ours}     & 3D    & {\color[HTML]{FE0000} 99.27}  & {\color[HTML]{FE0000} 94.02} & {\color[HTML]{FE0000} 97.80} & 8   \\ \hline
\end{tabular}
\end{table}

\subsubsection{KITTI 3D Pedestrian validation set benchmark}
In Table \ref{t:kitti_val}, we present a summary of the pedestrian tracking results. Our proposed system demonstrates superior performance compared to other modern tracking algorithms. Across all metrics, with the exception of ID switches (IDs), our method achieves the best results.
It's worth noting that AB3DMOT shows the lowest number of ID switches. However, this is primarily due to its significantly lower detection capability. To provide a fair comparison, we modified AB3DMOT to use the same detector as our SpbTracker, denoted as AB3DMOT*. With this modification, AB3DMOT* exhibits the highest number of ID switches.

\begin{table}[t]
\centering
\renewcommand{\arraystretch}{1.4} 
\caption{Custom Office Dataset, 3D person validation set benchmark}
\label{t:office_val}
\begin{tabular}{llllll}
\hline
\textbf{Model}    & \textbf{Input} & \textbf{sAMOTA}$\uparrow$ & \textbf{MOTA}$\uparrow$  & \textbf{recall}$\uparrow$  & \textbf{IDs}$\downarrow$ \\ \hline
PolarMOT & 3D    & 89.12  & 85.48 & 89.16 & 52   \\
AB3MOT*  & 3D    & 88.42  & 85.77 & 90.11 & 88   \\ \hline
\textbf{ours}     & 3D    & {\color[HTML]{FE0000} 92.12}  & {\color[HTML]{FE0000} 88.70} & {\color[HTML]{FE0000} 91.98} & {\color[HTML]{FE0000} 48}    \\ \hline
\end{tabular}
\end{table}
\subsubsection{Office 3D person tracking}
In Table \ref{t:office_val}, AB3DMOT* denotes the AB3DMOT algorithm using our detector instead of its original one. The office dataset presents a more challenging environment compared to KITTI due to the presence of crowded objects. Additionally, the LiDAR sensor used in our office dataset has a low range than KITTI's sensor and the frame rate is lower.
These factors contribute to the overall lower performance results compared to those obtained on the KITTI dataset. It's important to note that all tracking algorithm parameters, including those for our method and the other algorithms evaluated, remain consistent with the settings used in the KITTI experiments.
This comparison highlights the robustness of our tracking system across different environmental conditions and sensor configurations, while also demonstrating the challenges posed by more complex scenarios such as crowded indoor environments.

\subsubsection{KITTI 2D Pedestrian Tracking}
In Table \ref{t:test-benchmark}, we present the 2D Pedestrian tracking results obtained on the KITTI test set. Although our tracking is performed in 3D space, we can report 2D tracking results by projecting the 3D bounding boxes onto the image plane using camera intrinsic and extrinsic parameters. We then report the minimal axis-aligned 2D bounding boxes that fully enclose these projections as the tracks' 2D positions.
Despite focusing on 3D tracking and using 2D detections only as a secondary cue, our method achieves state-of-the-art results in terms of sMOTA, MOTA, MOTP, and IDSW metrics for the 3D modality(except for models considering only 2D image space). Notably, our sMOTA and IDSW metrics are state-of-the-art even when compared to 2D+3D multi-modal models.
This performance demonstrates the effectiveness of our 3D tracking approach, which can compete with and even surpass methods that directly operate in 2D or use both 2D and 3D information. Our results highlight the potential of 3D-focused tracking methods in achieving high performance across both 3D and 2D evaluation metrics.

\section{Conclusion}
In this paper, we analyzed the "TBD" 3D tracking algorithm and proposed several enhancements for improved person tracking. Our contributions include a person-biased detector, MCIoU, feature similarity measures, a specific person motion model, a robust filter, and long-term life-cycle memory, all of which lead to significant performance improvements.
However, the use of large-scale memory in the life-cycle model results in increased matching time. To address this, we optimized the code using C++, but further research is needed to develop a more compact algorithm. Future work will explore learning-based methods for end-to-end tracking with multi-modal, moving away from the rule-based TBD approach.

%%%%%%%%%%%%%%%%%%%%%%%%%%%%%%%%%%%%%%%%%%%%%%%%%%%%%%%%%%%%%%%%%%%%%%%%%%%%%%%%%%%%%%%%%%%%%%%%

\newpage
\bibliographystyle{splncs04}
\bibliography{ref}





% LNCS Homepage, \url{http://www.springer.com/lncs}, last accessed 2023/10/25
% \end{thebibliography}
\end{document}